\pdfoutput=1

\documentclass[11pt]{article}

\usepackage[final]{coling}

\usepackage{times}
\usepackage{latexsym}

\usepackage[T1]{fontenc}

\usepackage[utf8]{inputenc}

\usepackage{microtype}

\usepackage{inconsolata}

\usepackage{graphicx}
\newcommand{\added}[1]{{#1}}
\graphicspath{ {./figures/} }

\usepackage{hyperref}
\usepackage{float}
\usepackage{verbatim} 
\usepackage{apalike}
\restylefloat{figure}
\restylefloat{table}

\usepackage{booktabs}
\usepackage{multirow}
\usepackage{caption}
\usepackage{subcaption}
\usepackage{tgbonum}
\usepackage{tikz}
\usepackage{lscape}
\usepackage{csquotes}
\usepackage{textcomp}
\usepackage[force]{covington}
\usepackage{tipa}
\usepackage{mathptmx}
\usepackage{setspace}
\usepackage{adjustbox}

\title{Small Language Models can Outperform Humans in Short Creative Writing: A Study Comparing SLMs with Humans and LLMs}

\author{
 \textbf{Guillermo Marco\textsuperscript{\dag}},
 \textbf{Luz Rello\textsuperscript{\ddag}},
 \textbf{Julio Gonzalo\textsuperscript{\dag}}
\\
\\
 \textsuperscript{\dag}UNED, Madrid, Spain
 \\
 \textsuperscript{\ddag}IE University, Madrid, Spain
\\
 \small{
   \textbf{Correspondence:} \href{gmarco@lsi.uned.es}{gmarco@lsi.uned.es}
 }
}

\begin{document}
\maketitle
\begin{abstract}
In this paper, we evaluate the creative fiction writing abilities of a fine-tuned small language model (SLM), BART-large, and compare its performance to human writers and two large language models (LLMs): GPT-3.5 and GPT-4o. Our evaluation consists of two experiments: (i) a human study in which 68 participants rated short stories from humans and the SLM on grammaticality, relevance, creativity, and attractiveness, and (ii) a qualitative linguistic analysis examining the textual characteristics of stories produced by each model. In the first experiment, BART-large outscored average human writers overall (2.11 vs. 1.85), a 14\% relative improvement, though the slight human advantage in creativity was not statistically significant. In the second experiment, qualitative analysis showed that while GPT-4o demonstrated near-perfect coherence and used less cliche phrases, it tended to produce more predictable language, with only 3\% of its synopses featuring surprising associations (compared to 15\% for BART). These findings highlight how model size and fine-tuning influence the balance between creativity, fluency, and coherence in creative writing tasks, and demonstrate that smaller models can, in certain contexts, rival both humans and larger models.\footnote{All annotated data and model weights are available on GitHub: \url{https://github.com/grmarco/slm-creativity}.}
\end{abstract}

\section{Introduction}

The field of Natural Language Processing (NLP) has seen significant advancements due to Large Language Models (LLMs), which have demonstrated remarkable performance across a variety of tasks, including creative writing \cite{chang2024survey,wang2024weaver}. These models, such as GPT-3.5 and GPT-4 \cite{ouyang2022training,achiam2023gpt}, can generate human-like text and have set a new standard for many language-related applications. However, they require immense computational resources and large datasets, making them resource-intensive. While these large models have excellent performance, recent research has suggested that smaller models, or Small Language Models (SLMs), can achieve competitive results in certain tasks with a much lower computational cost \cite{schick2020s,bilenko_introducing_2024,chen_what_2024}. This leads to our core research question: \textbf{\textit{RQ0: Can a fine-tuned small language model be competitive in the field of creative, literary writing?}}

This study explores the capabilities of BART-large \cite{lewis_bart_2019}, an SLM, in a creative writing task: generating movie synopses based on a given title. The first goal is to compare the quality of these synopses to those written by humans. To explore this, we conducted a comprehensive study in which we collected over 24,000 manual assessments. Specifically, 68 participants evaluated 60 synopses across five dimensions —readability, understandability, relevance, attractiveness, and creativity— in three different experimental settings. With this dataset, we answer the central question: \textbf{\textit{RQ1: How do creative texts produced by SLMs compare with human equivalents in terms of readability, understandability, relevance, attractiveness, and creativity?}} Our findings show that, surprisingly, BART-large outperformed humans across all quality dimensions except creativity.

To further explore the role of human perception in evaluating AI-generated text, the experiment was conducted in three variants: (1) the readers were unaware of who wrote the text, (2) the readers were explicitly told whether the text was written by a human or a AI, and (3) the readers were told all texts were AI-generated, regardless of their true origin. This experimental setup aims to answer the question: \textit{\textbf{RQ2: How relevant are reader biases regarding the author's identity (human or AI) in their evaluation scores?}} The results demonstrated that knowing the author influenced how readers perceived the quality of the text, with AI-generated texts rated lower when their origin was revealed. This suggests that biases about AI authorship can negatively affect creativity assessments.

Since we are asking about creativity and this is a subjective attribute, we use the experiment to answer the question: \textit{\textbf{RQ3: How do readers assess creativity, and how does it correlate with other aspects of the texts?}}

Having conducted the experiments with human evaluators comparing texts written by an SLM and texts written by human authors, we arrive at our second goal of our research: to study the impact of model size on creative tasks. Thus, we arrive at the last research question: \textbf{\textit{RQ4: Do larger language models directly lead to more creative texts?}} For this purpose, we conducted a qualitative linguistic analysis that zooms in on the weak and strong points of each model regarding the generation of creative texts. The results of the analysis indicate that, while larger models such as GPT-4o produce more consistent and coherent text, they also tend to follow more predictable, formulaic patterns. In contrast, BART-large, despite being a smaller model, generated more novel and surprising content. Specifically, 15\% of the synopses produced by BART-large were deemed creatively novel, compared to only 3\% for GPT-4. This suggests that larger models reproduce natural language patterns more accurately, achieving greater consistency and fluency but often generating more predictable and formulaic outputs. In contrast, smaller models like BART-large, since they do not replicate these patterns as precisely, are more likely to deviate from predictable structures, leading to greater flexibility and originality. This highlights how the inherent limitations of smaller models can sometimes foster more creative and unexpected outputs.

This research highlights the potential of smaller models for creative tasks, demonstrating that SLMs can sometimes be competitive with larger models in areas like creativity. This opens up new possibilities for more efficient, task-specific models that balance the strengths of both large and small models.

\section{Related Work}

Early research has shown that small language models (SLMs) can perform well in creative text generation. For example, \citet{ravi2024humor} found that smaller models can produce humor when given proper guidance. Similarly, \citet{schick_its_2021} demonstrated that SLMs work effectively as few-shot learners with prompt-based techniques. \citet{eldan_tinystories_2023} introduced TinyStories, showing that models with fewer than 10 million parameters can still generate coherent short narratives, challenging the belief that larger models are always needed for fluent and coherent output. For a more detailed overview of small models, see \citet{chen_what_2024}.

Work on how model size and perceived authorship influence reader evaluations has produced mixed results. In poetry, \citet{porter2024ai} found that a large language model can sometimes be preferred over human-authored poems, especially if readers are unaware of the poem’s origin. In contrast, \citet{marco_pron_2024}, who compared a leading LLM (GPT-4) with a renowned human author and used expert literary critics as evaluators, concluded that the top human author was clearly favored, particularly for originality and literary quality. These findings suggest that preferences depend on the evaluator’s expertise, the model’s scale, and the author’s standing.

Our study explores a different combination: we compare a small language model (BART-large) against average human writers, evaluated by general readers. By varying the information provided about the text’s origin—explicitly stating that it is AI-generated, stating that it is human-written, or withholding the information—we investigate whether a smaller model can match or surpass the performance of both larger models and human authors under these conditions.

Assessing creativity itself remains a complex challenge. Tests like the Torrance Test of Creative Writing \cite{chakrabarty_art_2024} show that LLM-generated texts often do not meet the creative standards set by professional authors. For more information on creativity in machine learning, see \citet{franceschelli_creativity_2022}. In line with these findings, our study considers how small models fit into this broader landscape, examining how model size, author expertise, and audience perception interact.

\section{Experimental Design}
\label{sec:design}

To answer our research questions, we have designed two different experiments. The first one is an evaluation with humans that tries to quantify how far the texts written by our SLM are from those written by humans. The second experiment consists of qualitatively analyzing the linguistic similarities and differences between the SLM texts and the most popular LLMs right now: GPT-3.5\footnote{In July 2024, OpenAI announced that it would replace gpt-3.5 with gpt-4o-mini in the ChatGPT web version. By then this research had already been completed. However, we decided to keep the results of GPT-3.5 because it is still available for access through the API, but, above all, because of its proximity in performance with open source models as Mixtral \cite{jiang2024mixtral} or Gemma \cite{team2024gemma} that are still widely used.} and GPT-4o. The task consists of, given a potential movie title, writing an imaginary synopsis for a movie with that title. 
 
\subsection{SLM vs. Humans Methodology}
\label{sec:slm-vs-human}

\paragraph{The Small Language Model} For our main experiment we used the default BART-large pre-trained model \citep{lewis_bart_2019}. Recent work in humor generation \cite{ravi2024humor} shows that BART remains a strong model for creative writing tasks. We fine-tuned it for synopsis generation from titles. We rely on sampling and beam search with the default BART HuggingFace configuration\footnote{\url{https://huggingface.co/facebook/bart-large/blob/main/config.json}}, without adjusting the temperature parameter. Although it is often suggested that increasing temperature could foster greater creative variability, recent research \cite{peeperkorn2024temperature} indicates that higher temperatures mainly introduce randomness rather than yielding truly more creative outputs. Consequently, we maintain a fixed temperature setting. This ensures that observed differences in textual quality or reader perception stem from the model’s inherent capabilities and our experimental conditions rather than from artificial variability.

\paragraph{Dataset} The dataset for this study was created by merging the Corpus of Movie Plot Synopses with Tags (MPST) \citep{kar_mpst_2018}, the CMU Movie Summary Corpus \citep{bamman_learning_2013}, and the Wikipedia Movie Plots datasets \citep{noauthor_wikipedia_nodate}. Duplicate titles and synopses exceeding 1,024 tokens were removed to comply with the BART input limit. The dataset was split into 80\% training, 10\% validation, and 10\% test sets, with 42,049 examples for training and 5,257 for both validation and testing. Items for evaluation, including human-written and model-generated synopses, were randomly selected from the test set. To avoid evaluation biases, we restricted the selection to model/human synopses pairs of similar lengths (within ±15 tokens), and to movie titles with fewer than 1,000 votes on IMDb, to minimize the chance that assessors would recognize them. The average synopsis length was 79 tokens for model-generated and 78 for human-written texts.

\paragraph{Quiz design} We ask the assessor to evaluate a number of quality aspects of the text. We have run three experiments with a similar setup (see "Variants" below). For each experiment, we had two quizzes, each consisting of 60 synopses: half (30) human-written, and half (30) generated by our system. We decided to use this number of synopses because a sample size of 30 is large enough for statistical significance tests, and over 60 synopses would cause fatigue in our assessors. The synopses with human-written synopses in quiz A have model synopses in quiz B, and vice versa. Half of the participants were randomly assigned to quiz A and half were randomly assigned to quiz B. The synopses are displayed in random order for each subject.

Each of the 60 questions of a quiz asked the assessor to read a title and a synopsis. Then, the assessor is asked to evaluate, in a Likert scale of 0 to 4 (0 - not at all, 1- a little, 2- enough, 3 - a lot, 4 - completely) several key aspects: \textit{readability}, assessing whether the writing was grammatically correct; \textit{understandability}, determining if the synopsis made sense; \textit{relevance}, evaluating the connection between the synopsis and the title; \textit{informativity}, measuring how much information the synopsis provided about the film, including its genre (e.g., children's, romance, adventure, science fiction, crime, etc.); \textit{attractiveness}, judging whether the synopsis made the reader want to see the film; and \textit{creativity}. A final control question checked if the participant was already familiar with the title or any of the information in the synopsis.

Assessors were instructed not to search for any information about the movies before completing their assessments. After the quiz, they were asked to reflect on the factors they considered when evaluating creativity, on whether their evaluation criteria changed during the process, and to provide feedback on their overall experience.

\paragraph{Variants} In the main experiment, assessors were informed beforehand that the synopses were written by either humans or a computer system but were not told which specific synopsis belonged to whom during the evaluation, aiming to minimize bias. To assess how expectations about authorship might influence their judgments, two additional experiments were conducted: (1) in the "Revealed" variant, assessors were explicitly told the authorship of each synopsis; (2) in the "AI only" variant, assessors were misled to believe all synopses were generated by AI, even though half were human-written.
 
The 60 synopses were the same in the three experiments. Note that this is a between-subjects experiment design. A within-subject study would require that the two synopsis for the same title (the original, human-made synopsis and the one invented by the SLM taking the title as input) were evaluated by the same assessor. But if the assessor reads both synopsis, there are biases that might interfere. In particular, assessors might try to deduce which of the two alternative synopsis is AI-made, and their presuppositions might affect their scores. In fact, one of the experiments described in the paper confirmed that the scores assigned by our subjects changed when we gave them information about the authors of each synopsis.

\paragraph{Human Writers} The synopses used in this study corresponds to actual movies and were selected from various publicly available sources, primarily Wikipedia and movie databases. Given the nature of these sources, the human-authored synopses can be considered representative of average fiction writers (neither top fiction writers nor random humans).  

\paragraph{Assessors} We recruited 68 volunteer participants, all students in an international MBA master's program. While homogeneous in terms of age and educational level, they came from diverse cultural and academic backgrounds, including sciences, social sciences, and humanities.  Further details are provided in Appendix \ref{appx:assesors}.

\subsection{SLM vs. GPT-3.5 vs. GPT-4o Methodology}
\label{sec:slm-vs-chatgpt}

We designed a methodology to compare the texts generated by our SLM vs GPT-3.5 and GPT-4o in order to (i) reach a better, qualitative understanding of our quantitative results, and (ii) compare the results of the SLM and GPT-3.5 and GPT-4o (a large, state of the art model following prompts without fine-tuning, i.e., in zero-shot mode). 

In order to do this, we have first asked GPT-3.5 and GPT-4o to generate a synopsis for each of the titles in our dataset, simply using the prompt ``Invent a synopsis for a movie entitled [MOVIE TITLE]''. Then, we performed a comparative linguistic study of the synopses generated by the three LLMs. Our results shed some light on the comparative capabilities of small-size, fined tuned model and state-of-the-art models in zero-shot mode. 

First, we have manually examined all AI-generated synopses (BART, GPT-3.5 and GPT-4o), and conclude that there are five salient properties that are typical of AI-generated synopses (See Table \ref{tab:qualitative}): \textit{repetitiveness}, \textit{recurrent themes}, \textit{coherence with external facts}, \textit{internal coherence} and \textit{surprising associations}. Then, each synopsis has been analyzed and annotated by a linguist, taking into account the proposed properties. This provides a qualitative characterization of BART and GPTs synopses. 

\section{Results of SLM vs. Humans Experiment}
\label{sec:results}
Table \ref{tab:variants} shows the average scores for each quality aspect of the human vs. model-generated synopses, along with the statistical significance of the differences found. The table allows us to answer our first two Research Questions.


\subsection{RQ1: How do creative texts produced by SLMs compare with human equivalents in terms of {\em readability}, {\em understandability}, {\em relevance}, {\em attractiveness} and {\em creativity}?}

\begin{table*}
\footnotesize
\centering
\caption{Means and differences for each experiment (* denotes significant difference in Wilcoxon signed-rank test: **** for $p < .0001$, *** for $p < .001$, ** for $p < .01$, and * for $p < .05$ ) \label{tab:variants}}
\begin{tabular}{crlccc}
\hline
\multicolumn{6}{c}{\textbf{Means for each experiment}}                                                                                                                                                                                                       \\ \hline
                     & \multicolumn{1}{c}{\textbf{Aspects}}         & \multicolumn{1}{c}{\textbf{Author}}              & \textbf{Main}                              & \textbf{Revealed (-Main )}                & \textbf{``AI only'' (-Main)}            \\ \hline
                     &                                              & \textit{SLM}                                 & 3.035                                      & 2.695 (-11\%***)                          & 2.779 (-8\%**)                             \\
                     &                                              & \textit{human}                                   & 2.490                                      & 2.417 (-3\%)                              & 2.337 (-6\%*)                              \\
\multirow{-3}{*}{1}  & \multirow{-3}{*}{\textit{readability}}       & \textit{SLM - human} & \textbf{+22\%****} & \textbf{+12\%***} & \textbf{+19\%****} \\ \hline
                     &                                              & \textit{BART model}                              & 2.489                                      & 2.220 (-11\%*)                            & 2.323 (-7\%*)                              \\
                     &                                              & \textit{human}                                   & 2.125                                      & 2.085 (-2\%)                              & 2.016 (-5\%)                               \\
\multirow{-3}{*}{2}  & \multirow{-3}{*}{\textit{understandability}} & \textit{SLM - human}  & \textbf{+17\%**}   & +6\%              & \textbf{+15\%**}  \\ \hline
                     &                                              & \textit{BART model}                              & 2.308                                      & 2.153 (-7\%)                              & 2.296 (-1\%)                               \\
                     &                                              & \textit{human}                                   & 1.884                                      & 1.865 (-1\%)                              & 1.898 (+1\%)                               \\
\multirow{-3}{*}{3}  & \multirow{-3}{*}{\textit{relevance}}         & \textit{SLM - human} & \textbf{+23\%**}   & \textbf{+15\%***} & \textbf{+21\%***}  \\ \hline
                     &                                              & \textit{SLM}                                 & 2.159                                      & 2.067(-4\%)                               & 2.113 (-2\%)                               \\
                     &                                              & \textit{human}                                   & 1.941                                      & 1.962 (+1\%)                              & 2.024 (+4\%)                               \\
\multirow{-3}{*}{4}  & \multirow{-3}{*}{\textit{informativity}}     & \textit{SLM - human} & \textbf{+11\%*}    & +5\%              & +4\%               \\ \hline
                     &                                              & \textit{SLM}                                 & 1.440                                      & 1.365 (-5\%)                              & 1.675 (+16\%****)                          \\
                     &                                              & \textit{human}                                   & 1.221                                      & 1.392 (+14\%*)                            & 1.543 (+26\%****)                          \\
\multirow{-3}{*}{5}  & \multirow{-3}{*}{\textit{attractiveness}}    & \textit{SLM - human} & \textbf{+18\%**}   & -2\%              & +9\%               \\ \hline
                     &                                              & \textit{SLM}                                 & 1.413                                      & 1.462 (+3\%)                              & 1.717 (+22\%****)                          \\
                     &                                              & \textit{human}                                   & 1.459                                      & 1.530 (+5\%)                              & 1.762 (+21\%****)                          \\
\multirow{-3}{*}{6}  & \multirow{-3}{*}{\textit{creativity}}        & \textit{SLM - human} & -3\%               & -4\%              & -3\%               \\ \hline
\multicolumn{1}{l}{} & \multicolumn{1}{c}{}                         & \textit{SLM}                                 & 2.110                                      & 1.983 (-6\%*)                             & 2.146 (+2\%)                               \\
\multicolumn{1}{l}{} & \multicolumn{1}{c}{GLOBAL}                   & \textit{human}                                   & 1.853                                      & 1.875 (+1\%)                              & 1.930 (+4\%)                               \\
\multicolumn{1}{l}{} & \multicolumn{1}{l}{}                         & \textit{SLM - human} & \textbf{+14\%****} & \textbf{6\%*}     & \textbf{11\%***}   \\ \hline
\end{tabular}
\end{table*}


The results show that SLM-generated synopses outperform human-written ones in all aspects except creativity. The overall score for AI-generated synopses was 2.11 compared to 1.85 for human synopses (a 14\% improvement). Specifically, SLM synopses were 22\% more readable, 17\% more understandable, 23\% more relevant to the title, 11\% more informative, and 18\% more attractive, with all these differences being statistically significant. However, the SLM scored 3\% lower in creativity, though the difference was not statistically significant. 

Does the apprentice beat the master? The fact that a SLM, trained to mimic human-written texts, can outperform humans in such a creative task is notable, as it suggests that they are capable of producing better-than-human outputs even by merely imitating human examples. A possible explanation for the higher average score of the SLM could be that, in the learning process, generalization helps the SLM avoid gross mistakes, and therefore produce more homogeneous text. Then, it would get better averages just by avoiding lowest scores. However, this hypothesis is incorrect: although the standard deviation of humans in overall score is higher than in the SLM (0.461 vs. 0.355) -- which is consistent with the hypothesis -- the SLM actually gets more overall 5's (best possible score) than humans: not only the average is better, but the SLM writes more high quality synopses than the humans. For a more in-depth statistical analysis see Appendix \ref{appx:stats}.

\subsection{RQ2: How relevant are the expectations and biases of the reader with respect to AI vs. human authors?}

In the main experiment, assessors were unaware of the author of each text. To explore potential biases, we ran two additional experiments: one where the author (human or AI) was revealed, and another where readers were told that all synopses were AI-generated (although actually half of them were human-written). Results of these two additional experiments are also displayed in Table \ref{tab:variants}.

In the "revealed author" variant, the bias against AI-generated synopses becomes evident. The overall score difference between humans and AI decreased from 0.257 to 0.107, with AI synopses penalized by 6\% when the author was revealed. AI synopses scored lower in readability, understandability, and attractiveness, while human scores remained relatively stable. This suggests a clear negative bias against AI-generated content when directly compared to human authors, as the explicit framing highlights AI as a competitor.

In contrast, in the "AI-only" variant, where assessors believed all synopses were AI-generated, the dynamic changes. Both human and AI synopses were penalized in readability and understandability, likely due to preconceived notions about AI's limitations. However, attractiveness ratings increased for both groups, and some assessors expressed surprise or admiration for the perceived quality of the synopses. This "wow" effect may stem from the novelty or unexpected competence of AI in generating coherent and engaging content, especially when readers are not directly comparing it to human output. Interestingly, this effect suggests that AI's perceived performance can benefit when it is not framed as a direct competitor to human creativity.

These findings indicate that revealing the author introduces a bias against AI-generated content, lowering reader evaluations, particularly in direct comparisons with human authors. However, when AI is presented as the sole source of content, some biases may transform into a positive re-evaluation of its capabilities, highlighting the nuanced interplay between reader expectations and the framing of authorship.

\subsection{RQ3: How do readers assess creativity, and how it correlates with other aspects of the texts?}

We deliberately chose not to provide readers with a specific definition of creativity, to avoid introducing our own biases into their evaluations. Our readers were free to interpret the term "creativity'', following \cite{colton_computational_2012}: \textit{``Computer Creativity is the philosophy, science and engineering of computational systems which, by taking on particular responsibilities, exhibit behaviors that unbiased observers would deem to be creative.``} Once readers completed the evaluation, we asked them what factors in the synopses influenced their quantitative assessment of creativity. The reason is that we are focused on whether SLM texts give the appearance of creativity to readers in general (something that is measurable), not whether they are truly creative from an artist's perspective. After completing the experiment, participants were asked what factors influenced their creativity scores.

The responses fell into two main categories: (1) \textit{Writing style}: 21 out of 68 assessors noted that they judged creativity based on elements like cohesion, vocabulary, and the overall fluency of the text. (2) \textit{Originality and predictability}: Another 21 assessors mentioned that they assessed creativity by considering how familiar or predictable the storyline seemed. A few assessors offered additional insights, such as creativity being linked to how detailed the synopsis was or how well the synopsis aligned with their expectations based on the title. Interestingly, 22 assessors did not answer the question regarding creativity.

\paragraph{Correlation of Creativity with Other Quality Aspects}

\begin{table}
\scriptsize
\centering
\caption{Spearman's correlation for each aspect. (* denotes significant correlation: ** for $p < .01$, * for $p < .05$)}\label{tab:correlations}
\setlength{\tabcolsep}{3pt} 
\begin{tabular}{rlllll}
  \hline
 & \textit{\shortstack{readability}} & \textit{\shortstack{understand- \\ ability}} & \textit{\shortstack{attracti- \\ veness}} & \textit{\shortstack{infor- \\ mativity}} & \textit{\shortstack{relevance}} \\ 
  \hline
readability &  &  &  &  &  \\ 
understandability &  0.71** &  &  &  &  \\ 
attractiveness &  0.17     &  0.30*    &  &  &  \\ 
informativity &  0.42*   &  0.66**  &  0.42*   &  &  \\ 
relevance &  0.66**  &  0.70**  &  0.28*    &  0.61**  &  \\ 
creativity &  0.42*   &  0.41*    &  0.25     &  0.60**  &  0.69** \\ 
   \hline
\end{tabular}

\end{table}

We further investigated how creativity correlates with other quality aspects like relevance, informativeness, and attractiveness by calculating the correlations between these aspects for each synopsis. Table \ref{tab:correlations} shows that creativity has a strong positive correlation with relevance (0.69) and informativeness (0.60). This suggests that readers often perceive a more creative text when it is more relevant to the title or provides more detailed information. For instance, one reader noted that more detailed synopses appeared more creative, which aligns with the observed correlation between creativity and informativeness.

Relevance also showed strong correlations with all other aspects except attractiveness, indicating that synopses closely aligned with the title were generally rated higher across all dimensions. On the other hand, attractiveness was the least correlated with other aspects, with a correlation below 0.3 for all aspects except informativeness. This suggests that attractiveness is likely influenced by personal preferences or prior movie-watching experiences, making it a more subjective metric.

It’s also noteworthy that creativity and attractiveness were only weakly correlated (0.25), indicating that creativity doesn’t necessarily drive the appeal of a synopsis. This could help explain why blockbuster movies often repeat successful clichés—while they may not be seen as highly creative, they can still be considered attractive or appealing to a broad audience.

\section{SLM vs LLMs: Results of the Linguistic Analysis}

Up to this point, we have focused on the comparison between human-generated texts and those produced by BART, where BART demonstrated superior performance in most evaluated dimensions. Having established this, we now turn our attention to a more detailed examination of BART in contrast to larger language models (LLMs), such as GPT-3.5 and GPT-4o. This analysis will explore the distinct linguistic features of these models, with a particular emphasis on how the larger LLMs tend to surpass BART in various aspects, especially in the case of GPT-4o. For concrete examples of the linguistic features discussed here, see Appendices \ref{appx:examples} and \ref{appx:linguistic}.

\begin{table*}[t!]
\footnotesize
\added{
\caption{Frequencies of the qualitative salient linguistic properties of SLM-generated synopses. For the formulaic phrases, we report the percentage of synopses in which they appear and the total number of appearances in parentheses. *The average of the 25\% longer synopses is 141,8 words/text for BART, 133,9 words/text for GPT-3.5 and 89,9 words/text for ChatGPT 4o.}
\label{tab:qualitative}
\centering
\begin{tabular}{llrrr}
& & BART (SLM) & GPT-3.5 & GPT 4o \\ 
 & & (fine-tuned) & (zero-shot)  & (zero-shot) \\ 
\hline
{\bf Formulaic Phrases}     & With at least one clich\'e phrase &  83,3\% (50) & {\bf 100\% (60)}  & 26,7\% (16) \\ 
                            & With two or more clich\'e phrases & {\bf 46,7\% (28)} &  25,0\% (15) & 0\% \\                      
                    \hline
{\bf External Coherence}  & Overall coherence with external facts & 86,7\% & 91,7\%  & {\bf 100\%} \\ 
                    & When referring to a place and a date & 62,5\% & 60,0\% & {\bf 100\%} \\ 
                    & When referring to a specific historical event & 50,0\%  & 30,0\% & {\bf 100\%} \\ \hline
{\bf Internal Coherence}  & Overall & 68,3\% & 95,0\% & {\bf 100\%} \\ 
                    & Of the top 25\% longer synopses* & 33,3\% &  100\% & {\bf 100\%} \\ \hline
{\bf Surprising Associations} &  & {\bf 15,0\%} & 3,3\%  & 1,6\% \\  \hline
{\bf Recurrent Themes}   & Adventure         & 3,3\%       & 20,0\%   & {\bf 25,0\%} \\
                         & Family            & {\bf 13,3\%}      & {\bf 13,3\%}   & 8,3\% \\
                         & Friendship        & {\bf 11,7\%}      & 5,0\%    & 3,3\% \\                    
                         & Love              & {\bf 36,7\%}      & 18,3\%   & 11,7\% \\     
                         & Mystery \& Crime          & 18,3\%      & 16,7\%   & {\bf 28,3\%} \\  
                         & Personal Growth   & 3,3\%       & {\bf 23,3\%}   & 20,0\% \\   
                         & War               & {\bf 13,3\%}      & 3,3\%    & 3,3\%  \\                           
                         
\hline
\end{tabular}
}
\vspace{-0.3cm}
\end{table*}

\subsection{Repetitiveness and Formulaic Phrases}

Table \ref{tab:qualitative} shows that the synopses generated by transformers tend to be formulaic and rely on cliché phrases. This repetitiveness was measured based on the frequency of certain collocations and words, with common phrases such as “[It/The film] tells the story of” appearing in 75\% of BART synopses and 35 times in GPT-3.5 summaries. The most cliché prone model is GPT-3.5 (100\% of the synopsis have at least one cliché), followed by BART (83,3\%). GPT-4o is substantially better, with just one in four synopsis having a cliché phrase (26,7\%); but it repeats vocabulary across the synopses.

Both models display varying degrees of repetition, with BART having more than 83\% of its synopses containing at least one cliché. GPT-4o, however, avoids multiple clichés, but repeats vocabulary across different texts.

A small set of phrases dominates the texts generated by these models, contributing to a lack of perceived creativity. The most common phrases include “[It/The film] tells the story of” and “The film is set in,” leading to a certain repetitiveness. GPT-4o, although less reliant on cliché phrases, uses longer but equally formulaic sentences such as “life is turned upside down” or “love can sometimes be the wildest move of all.” This predictability limits the models' originality, even though human-generated synopses also exhibit formulaic tendencies.

In addition to clichés, frequent collocations like “car accident” and “successful businessman” further highlight the repetitiveness of both models. BART is less predictable in its use of clichés than GPT-3.5, which, even in zero-shot mode, shows a higher degree of predictability. However, GPT-4o introduces more variety with a growing use of proper names (88.3\% of its texts), less common words, and unique adjective-noun combinations, contributing to greater linguistic diversity.

\subsection{Recurrent Themes}

The analysis shows that both BART and GPT-3.5-generated synopses tend to focus on a narrow set of recurrent themes, which may contribute to a perceived lack of creativity. BART's synopses primarily revolve around love (38.3\%), crime (18.3\%), war (13.3\%), and family drama (11.7\%), with many love stories resembling soap operas. GPT-3.5 focuses on personal growth (23.2\%), adventure (20\%), love (18.3\%), and mystery (16.7\%). GPT-3.5’s synopses emphasize themes like self-discovery, the "true meaning" of love, and intimate human emotions more frequently than BART. 

GPT-4o also leans heavily into emotional narratives, particularly mysteries (28.3\%) and adventures (25\%), but introduces more original character types, such as archaeologists and chefs, compared to BART's simpler characters. While both models rely on a limited range of themes, similar to human-made synopses, GPT-3.5 tends to explicitly mention genres like drama and thriller more frequently. 

\subsection{Coherence with External Facts}

AI-generated synopses sometimes contain factual errors due to a lack of explicit world knowledge. Examples include setting the English Civil War in the 1920s or referencing the Iron Curtain after it had already fallen. However, most synopses (91.7\% from GPT-3.5 and 86.7\% from BART) are free of such errors, largely because they avoid referencing historical events or specific facts tied to dates and places.

When focusing on synopses that do mention external facts, both BART and GPT-3.5 show low reliability, with BART being accurate 62.5\% of the time regarding places and dates, and GPT-3.5 at 60\%. In references to specific historical events, BART is accurate 50\% of the time, while GPT-3.5 only reaches 30\%. GPT-4o, on the other hand, does not display any inconsistencies in these areas, but only 5\% of its synopses refer historical events. These inconsistencies do not heavily impact average manual assessments because the synopses rarely mention historical places and dates. 

Additionally, GPT-4o shows significant improvements in coherence, both in its handling of external facts and in maintaining overall semantic coherence compared to earlier models like GPT-3.5.

\subsection{Internal Coherence}

The analysis reveals that AI-generated synopses can struggle with internal coherence, despite their short length. For example, one BART synopsis describes a character's father dying in a car accident, only to later state that the father gets married. In total, 20 of BART’s synopses lack internal consistency (i.e. only 68.33\% are coherent). The issue worsens with longer synopses, where only 33.33\% of BART’s top 25\% longer synopses are coherent, indicating an inverse correlation between coherence and length.

In contrast, GPT-3.5 performs much better in terms of internal consistency, with 95\% of its synopses being coherent, and all of its longer synopses maintaining consistency. Additionally, all GPT-4o synopses are internally coherent. While internal coherence is distinct from creativity, GPT's larger models clearly outperform smaller models like BART, especially for longer generation tasks.

\subsection{Surprising Associations}

The analysis summarizes in Table \ref{tab:qualitative} reveals that AI-generated synopses sometimes produce surprising or humorous associations, which could be perceived as creative. BART exhibited such associations in 9 out of 60 synopses (15\%), whereas GPT-3.5 and GPT-4o displayed this only in two and one synopses, respectively. In our data, BART is four times more likely to generate unexpected content compared to the larger models.

It is remarkable that larger models such as GPT-3.5 and GPT-4o, despite producing more fluent, natural and consistent text, are less likely to produce creative or surprising associations, and this is a crucial consideration in automatic creative writing. Note that we see this effect in humans, too: children has less knowledge and skills, but are in average much more creative than adults. Gaining knowledge makes humans more functional, but also more predictable. 

\subsection{RQ4: Do larger, better language models directly lead to more creative texts?}

At this point, we are ready to answer our last research question.

Both BART and GPT-3.5 tend to rely on cliché phrases and recurrent topics, with BART being slightly more creative, as 15\% of its synopses contain surprising associations compared to only 2\% for GPT-3.5. However, GPT-3.5 and GPT-4o outperform BART in internal consistency (95\%-100\% vs. 68\%).

Although more research is needed to confirm this, our data suggests a trade-off between model size and creativity: larger models like GPT-3.5, trained for fluency and naturalness, tend to produce more consistent but less creative texts. This may be due to their predictability, which limits their ability to generate surprising content. Probably, the most sensible way to overcome this issue in collaborative human-AI writing is via prompting: an original, carefully designed prompt may result in a surprising outcome. Despite BART's limitations in coherence and creativity, its synopses were preferred by readers over the human-texts in all quality aspects except creativity, indicating that these limitations may not significantly affect overall manual assessments.

Comparing with human made synopses, BART limitations (coherence with external facts, internal coherence for larger texts, use of cliches and recurrent topics) are not significant enough to produce a lower manual assessment, as they are preferred by our set of readers in all quality aspects except creativity. If a small model can achieve these results, it is reasonable to expect that GPT-3.5 and GPT-4o would perform similarly or even better.

\section{Conclusions}

This study highlights the potential of small language models (SLMs) in creative writing tasks, particularly in generating short stories. Our experiments indicate that fine-tuned SLMs like BART-large are rated more favorably than average human writers by general readers in several aspects, including readability, understandability, relevance, and informativeness. However, when it comes to creativity, humans maintain a slight perceived edge, though the difference observed in our data is not statistically significant.

Interestingly, our findings suggest that it does not take a highly sophisticated large language model (LLM) to meet or even surpass human performance in many dimensions of creative writing, particularly when evaluated by general audiences. SLMs, despite their smaller size and lower computational requirements, demonstrated a strong ability to generate texts that resonated with readers, challenging assumptions about the necessity of larger models in this domain.

Our results also highlight the significant role of biases in shaping reader perceptions of AI-generated text. When AI is explicitly identified as the author, reader assessments tend to penalize its perceived creativity and overall quality. However, in scenarios where the AI is not competing directly against human authors or is evaluated anonymously, its output often receives more favorable ratings. This effect underlines how context and framing influence the reception of AI-generated content.

Comparing the SLM with larger language models (LLMs) ---GPT-3.5 and GPT-4o in our experiments--- our qualitative linguistic analysis revealed that while larger models produce more consistent and coherent texts, they often generate more predictable and formulaic narratives. GPT-4o, despite its near-perfect internal and external coherence, produced stories that were seen as novel only 3\% of the time. In contrast, the SLM generated novel content in 15\% of its stories, indicating a higher degree of creativity despite its smaller size. This suggests a trade-off between model size and creative flexibility, where larger models may prioritize consistency over originality.

These findings emphasize that SLMs can be competitive with both (average) humans and larger models in creative writing tasks, particularly when the task benefits from creative flexibility over strict consistency. It underscores that a Large Language Model is not always necessary, and that fine-tuned smaller models can be more suitable and efficient for specific applications. This opens up possibilities for developing more efficient, task-specific models that balance creativity, fluency, and coherence without the computational overhead of larger models.

\section*{Scope and Limitations of the Study}

SLMs have clearly received better average scores than humans in our experiment. Initially, we did not expect such a strong result. Although language models are trained with more text than a human can read in a lifetime, their generalization abilities are still weak in comparison with humans, and they do not connect language with real world knowledge. Our qualitative analysis, in fact, confirms the limitations of language models. What, then, is the correct interpretation of our results? Do they actually mean that SLM or LLMs can already perform certain creative tasks better than humans? In order to properly answer this question, let us review our experimental setting and how it constrains the scope of our results.

\paragraph{The task} 

As we have noted earlier, our task requires writing short texts, where internal coherence is less of a challenge. We chose it because it is well known that SLMs struggle to maintain coherence, and we wanted to measure their creative writing abilities without that factor influencing the results. We cannot, therefore, extrapolate our results to longer texts.

Also, producing a movie synopsis is a special kind of writing task. The synopses in the training set were not created from just a movie title, of course, but from the movie itself. Their goal is to spark the reader’s interest without revealing too much of the plot. It is a good choice for evaluation because it is a constrained task, and because there are many examples to learn from. But it is not a canonical creative writing task, and more work is needed before extrapolating our results to other tasks. 

\paragraph{The human writers we are comparing against} 
Although the set of movies used in our experiment with readers were chosen randomly, we had a few constraints. One of them is that the movie must have less than 1,000 reviews on IMDb, which reduces the possibility of our readers being familiarized with the actual movie (in order to avoid any biased evaluations). A side effect of this constraint is that less popular movies may correlate with lesser quality movies. We have checked our dataset with IMDb, and the average rating of the movies in the training set is 6.2 (with a median of 6.4); in the 60 movies chosen for the quiz, on the other hand, the average is 5.7 (median 5.9). The difference is not large, but it may be a partial explanation of why the SLM is able to apparently improve the synopses used in its learning process.

\paragraph{The Assessors: Quality as Popularity} 

Our assessors are not professional creative writers or critics; instead, we use a popularity criterion in our evaluation design ({\em Do target readers enjoy the synopses?}) instead of a professional criteria ({\em How do critics assess the synopsis?}). Both types of assessment are legitimate and complementary, and also often contradictory (for instance, blockbusters are usually much more appreciated by the audience than by the critics). In our experiment, we seemed to have reached (and surpassed) the threshold in which machines are able to match human writings in terms of popularity criteria; and this suggests that relying solely on popularity may no longer suffice for this type of evaluation. We may need to venture into the (turbulent) waters of how to properly assess the outputs of a creative process.

Overall, we believe that the superiority of SLM in our experiment is meaningful and has implications in the field, but it should not be overstated: we must extend the experimentation to more complex and naturalistic creative writing tasks, and we must go beyond popularity for a better understanding of the potential of SLM for this type of tasks. 

\section*{Acknowledgements}
This work has been financed by the European
Union (NextGenerationEU funds) through the “Plan
de Recuperación, Transformación y Resiliencia”, by
the Ministry of Economic Affairs and Digital Transformation and by UNED University. However,
the points of view and opinions expressed in this
document are solely those of the authors and do
not necessarily reflect those of the European Union
or European Commission. Neither the European
Union nor the European Commission can be considered responsible for them.

Guillermo Marco's work was funded by Spanish government Ph.D. research grant ({\it Ministerio de Universidades}) FPU20/07321 and a scholarship of the Madrid City Council for the Residencia de Estudiantes (Course 2023-2024).

\appendix

\section{Likert Distributions}
\label{appx:likert}
\begin{figure*}

     \centering
         \centering
        \resizebox{0.8\textwidth}{!}{\input{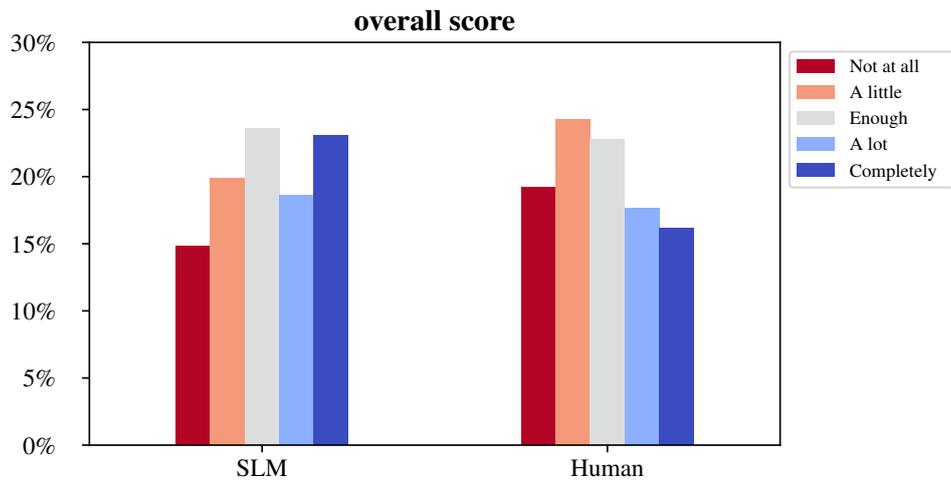}}
        \caption{Likert overall score distribution in the main experiment\label{fig:distribution-overall}}
\end{figure*}

\begin{figure*}
     \centering
         \centering
        \resizebox{0.8\textwidth}{!}{\input{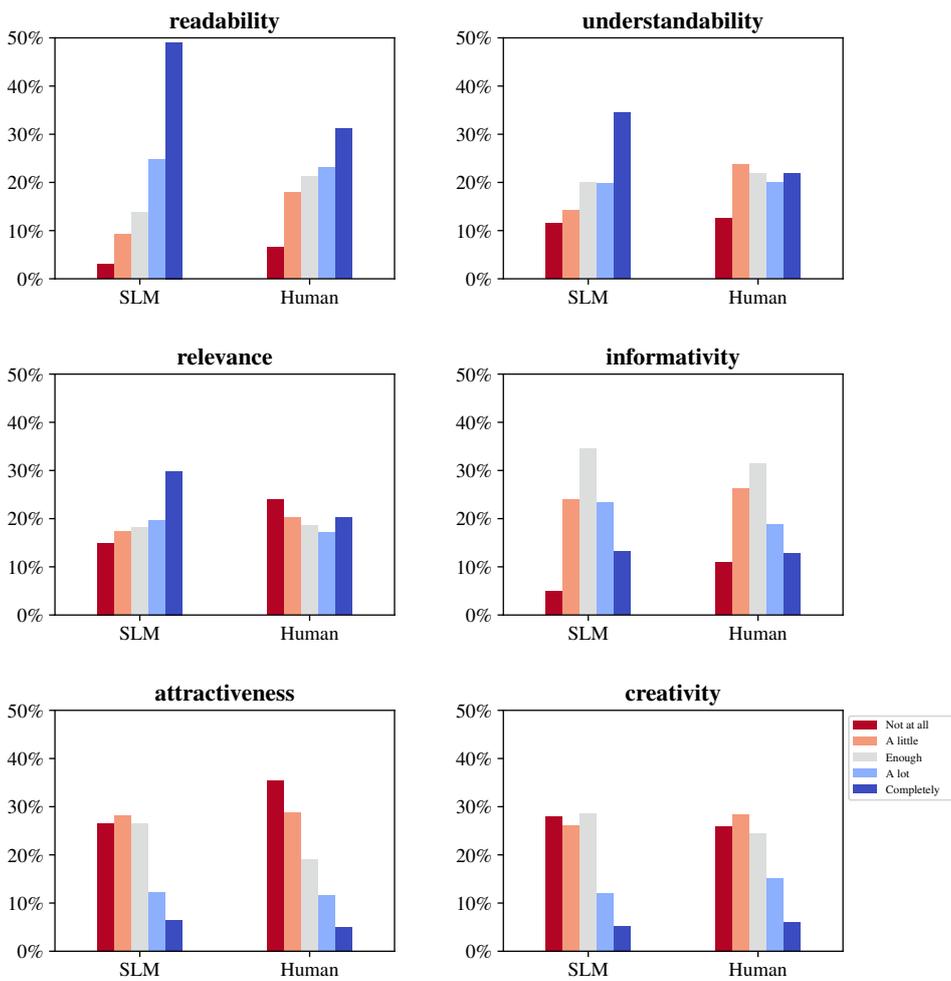}}
        \caption{Likert score distributions for each quality aspect in the main experiment.\label{fig:distributions}}
\end{figure*}

Figures \ref{fig:distribution-overall} and \ref{fig:distributions} show the distribution of scores on each item of the Likert scale (from 0 to 4) used in the SLM vs. humans experiment. 

\section{Assessors Information}
\label{appx:assesors}

\begin{table}
\small
\centering
\caption{Number of participants in each quiz. The titles with human written synopses in quiz A have SLM synopses in quiz B and vice versa.}
\label{tab:users}
\begin{tabular}{@{}l|cc@{}}
\toprule
\multicolumn{1}{c|}{\textbf{Experiment variant}} & \multicolumn{1}{c}{\textbf{Quiz A}} & \multicolumn{1}{c}{\textbf{Quiz B}} \\ \midrule
\textit{main experiment}                                  & 12                                  & 9                                   \\
\textit{``revealed'' variant}                                & 10                                  & 10                                  \\ 
\textit{``AI only'' variant}                             & 13                                  & 14                                  \\
\bottomrule
\end{tabular}
\end{table}

We recruited 68 volunteer participants for our study.
All subjects recruited for our experimentation were students of an international Master’s program in Management. All students were proficient in English (the official language of the Master’s program), although they come from different countries and therefore have different cultural backgrounds. They also have different academic backgrounds (sciences and technology, social sciences and humanities). Our set of assessors is, therefore, relatively homogeneous in terms of age and current educational level (all are Master’s students), but quite diverse in cultural and academic backgrounds. 

\begin{figure*}
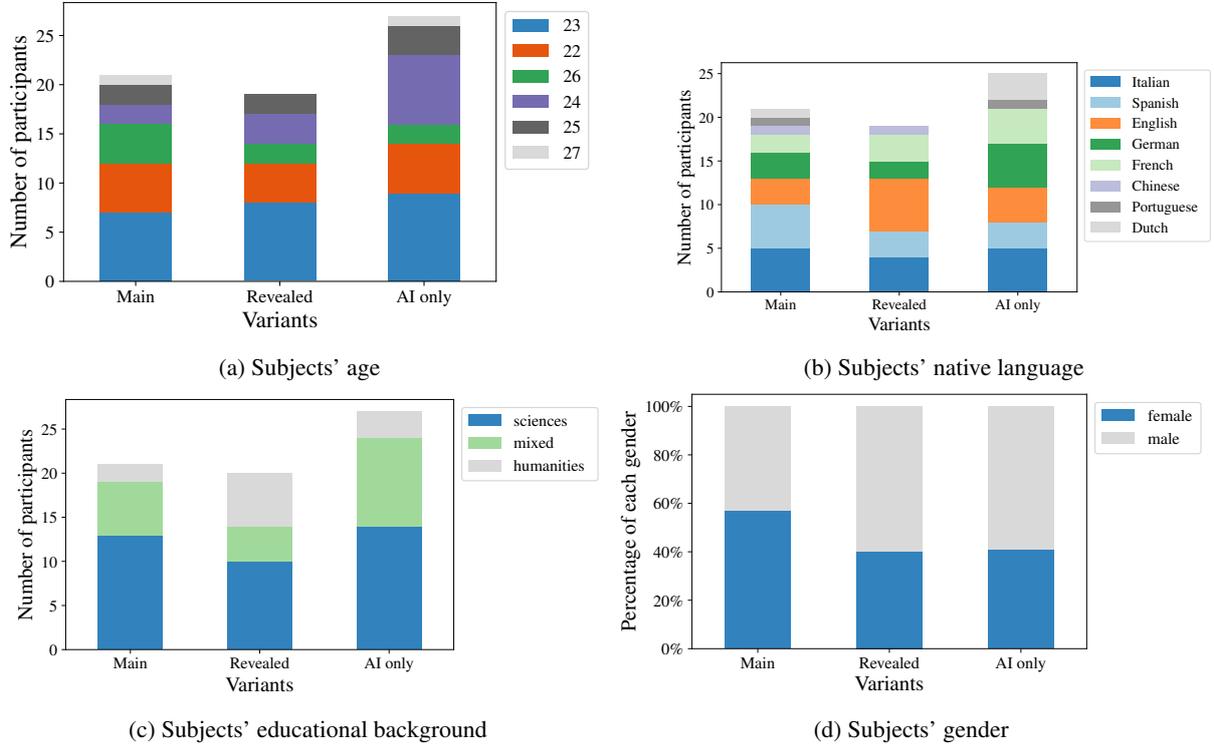

     \centering
     \begin{subfigure}[b]{0.49\textwidth}
         \centering
        \resizebox{1\textwidth}{!}{\input{figures/age-stack.pgf}}
        \caption{Participants’ Age Distribution}
     \end{subfigure}
     \hfill
     \begin{subfigure}[b]{0.45\textwidth}
         \centering
        \resizebox{1\textwidth}{!}{\input{figures/language-stack.pgf}}
         \caption{Subjects' native language}
     \end{subfigure}
     \hfill
     \begin{subfigure}[b]{0.49\textwidth}
         \centering
        \resizebox{1\textwidth}{!}{\input{figures/education-stack.pgf}}

         \caption{Subjects' educational background}
     \end{subfigure}
     \begin{subfigure}[b]{0.49\textwidth}
         \centering
        \resizebox{1\textwidth}{!}{\input{figures/gender-stack.pgf}}

         \caption{Subjects' gender}
     \end{subfigure}
        \caption{Participants' profile for each variant.}
        \label{fig:subjects}
\end{figure*}

Figure \ref{fig:subjects} summarizes the profile of the assessors in each of the three experiments carried out.

\section{Additional Statistical Analysis}
\label{appx:stats}

\subsection{Standard Deviations of the Main Experiment}

Table \ref{tab:bot-human} shows the means and standard deviations of the results of the main experiment.

\begin{table*}
\caption{Means of each aspect for human and SLM in the main experiment. (* denotes significant differences between human and SLM for this aspect in Wilcoxon signed-rank test: **** for $p < .0001$, *** for $p < .001$, ** for $p < .01$, and * for $p < .05$ ).}
\centering

\label{tab:bot-human}
\small
\begin{tabular}{crclccc}
\hline
\multicolumn{1}{l}{} & \multicolumn{1}{c}{\textbf{Properties}}     & \textbf{Questions}                                                                                                         & \textbf{Author}  & \multicolumn{1}{l}{\textbf{Mean}} & \multicolumn{1}{l}{\textbf{$SLM-human$}} & \multicolumn{1}{l}{\textbf{SD}} \\ \hline
\multirow{2}{*}{1}   & \multirow{2}{*}{\textit{readability}}       & \multirow{2}{*}{Is the writing grammatically correct?}                                                                     & \textit{SLM} & 3.035                             & \multirow{2}{*}{\textbf{(+22\%****)}}        & 0.506                                           \\
                     &                                             &                                                                                                                            & \textit{human}   & 2.49                              &                                              & 0.623                                           \\ \hline
\multirow{2}{*}{2}   & \multirow{2}{*}{\textit{understandability}} & \multirow{2}{*}{Does the synopsis make sense?}                                                                             & \textit{SLM} & 2.489                             & \multirow{2}{*}{\textbf{(+17\%**)}}          & 0.654                                           \\
                     &                                             &                                                                                                                            & \textit{human}   & 2.125                             &                                              & 0.684                                           \\ \hline
\multirow{2}{*}{3}   & \multirow{2}{*}{\textit{fidelity}}          & \multirow{2}{*}{Does the synopsis relate to the title?}                                                                    & \textit{SLM} & 2.308                             & \multirow{2}{*}{\textbf{(+23\%**)}}          & 0.826                                           \\
                     &                                             &                                                                                                                            & \textit{human}   & 1.884                             &                                              & 0.772                                           \\ \hline
\multirow{2}{*}{4}   & \multirow{2}{*}{\textit{informativity}}     & \multirow{2}{*}{\begin{tabular}[c]{@{}c@{}}How much information does \\ the synopsis provide about the film?\end{tabular}} & \textit{SLM} & 2.159                             & \multirow{2}{*}{\textbf{(+11\%*)}}           & 0.4                                             \\
                     &                                             &                                                                                                                            & \textit{human}   & 1.941                             &                                              & 0.578                                           \\ \hline
\multirow{2}{*}{5}   & \multirow{2}{*}{\textit{attractiveness}}    & \multirow{2}{*}{\begin{tabular}[c]{@{}c@{}}Does this synopsis make you want \\ to watch the film?\end{tabular}}            & \textit{SLM} & 1.440                             & \multirow{2}{*}{\textbf{(+18\%**)}}          & 0.456                                           \\
                     &                                             &                                                                                                                            & \textit{human}   & 1.221                             &                                              & 0.438                                           \\ \hline
\multirow{2}{*}{6}   & \multirow{2}{*}{\textit{creativity}}        & \multirow{2}{*}{Do you find the synopsis creative?}                                                                        & \textit{SLM} & 1.413                             & \multirow{2}{*}{(-3\%)}                      & 0.437                                           \\
                     &                                             &                                                                                                                            & \textit{human}   & 1.459                             &                                              & 0.435                                           \\ \hline
\multicolumn{1}{l}{} & \multicolumn{1}{c}{\multirow{2}{*}{GLOBAL}} & \multirow{2}{*}{\textbf{GLOBAL}}                                                                                           & \textit{SLM} & 2.110                             & \multirow{2}{*}{\textbf{(+14\%**)}}          & 0.355                                           \\
\multicolumn{1}{l}{} & \multicolumn{1}{c}{}                        &                                                                                                                            & \textit{human}   & 1.853                             &                                              & 0.461                                           \\ \hline
\end{tabular}
\end{table*}

\subsection{Linear Mixed-Effects Models}

Linear Mixed-Effects Models \citep{bates2014fitting,kaptein2016using} are a statistical method that consist of a linear regression where the model also includes random effects to account for the variance produced by the subjects and the set of the items (in our case, synopses) selected for the experiment. We have performed a Linear Mixed-Effects Models regression with two goals in mind. First, we want to verify the results we observed in the previous sections with a more in-depth statistical analysis. And, secondly, we want to check if the subject profiles influence their assessments. 

We have experimented with two different models. Both included the random effects of assessors, title of the synopsis (with 60 possible values) and their interaction with the 'writer' (human or SLM). As fixed effects we study those that allow us to check the reliability of our analysis: the effect of age, educational background (humanities or sciences), their native language and, mainly, the effect of the writer (human or SLM).

For each model, we checked that the error term is normally distributed and that there is no correlation between the model predictions and the residual. All models were computed with R {\texttt lmerTest} package \citep{kuznetsova2017lmertest} 


\begin{table*}
\footnotesize
\centering
\caption{Linear mixed-effects model for the model \
\textit{answer = writer * question 
                +  (1+writer|title) + (1+writer|subject)}}
\label{tab:model}
\begin{tabular}{lccccc}
\textbf{}                        & \multicolumn{5}{c}{\textbf{answer}}                                                                                                                           \\ \hline
\textit{Predictors}              & \textit{Estimates}            & \textit{std. Error}           & \textit{CI}                   & \textit{t-value}              & \textit{p}                    \\ \hline
(Intercept for human creativity) & 1.47                          & 0.12                          & 1.23 – 1.70                   & 12.02                         & \textbf{\textless{}0.001}     \\
SLM (with human creativity as origin)              & -0.06                         & 0.11                          & -0.28 – 0.15                  & -0.56                         & 0.578                         \\ \hline
readability                      & 1.07                          & 0.06                          & 0.94 – 1.19                   & 16.70                         & \textbf{\textless{}0.001}     \\
understandability                & 0.69                          & 0.06                          & 0.57 – 0.82                   & 10.82                         & \textbf{\textless{}0.001}     \\
relevance                        & 0.42                          & 0.06                          & 0.29 – 0.54                   & 6.50                          & \textbf{\textless{}0.001}     \\
informativity                    & 0.49                          & 0.06                          & 0.37 – 0.62                   & 7.69                          & \textbf{\textless{}0.001}     \\
attractiveness                   & -0.25                         & 0.06                          & -0.38 – -0.13                 & -3.92                         & \textbf{\textless{}0.001}     \\
SLM $\times$ readability               & 0.60                          & 0.09                          & 0.42 – 0.78                   & 6.65                          & \textbf{\textless{}0.001}     \\
SLM $\times$ understandability         & 0.42                          & 0.09                          & 0.24 – 0.60                   & 4.63                          & \textbf{\textless{}0.001}     \\
SLM $\times$ relevance                 & 0.49                          & 0.09                          & 0.31 – 0.66                   & 5.39                          & \textbf{\textless{}0.001}     \\
SLM $\times$ informativity             & 0.27                          & 0.09                          & 0.09 – 0.44                   & 2.93                          & \textbf{0.003}                \\
SLM $\times$ attractiveness            & 0.28                          & 0.09                          & 0.10 – 0.46                   & 3.07                          & \textbf{0.002}                \\
\textbf{Random Effects}          & \multicolumn{1}{l}{\textbf{}} & \multicolumn{1}{l}{\textbf{}} & \multicolumn{1}{l}{\textbf{}} & \multicolumn{1}{l}{\textbf{}} & \multicolumn{1}{l}{\textbf{}} \\ \hline
$\sigma^{2}$                            & \multicolumn{5}{l}{1.29}                                                                                                                                      \\
$\tau_{synopsis}$                        & \multicolumn{5}{l}{0.18}                                                                                                                                      \\
$\tau_{subject}$                         & \multicolumn{5}{l}{0.21}                                                                                                                                      \\
$\tau_{synopsis \times SLM}$               & \multicolumn{5}{l}{0.19}                                                                                                                                      \\
$\tau_{subject \times SLM}$               & \multicolumn{5}{l}{0.10}                                                                                                                                      \\
$N_{title}$                          & \multicolumn{5}{l}{60}                                                                                                                                        \\
$N_{subject}$                           & \multicolumn{5}{l}{21}                                                                                                                                        \\
Observations                     & \multicolumn{5}{l}{7560}                                                                                                                                      \\
Marginal $R^2$ / Conditional $R^2$     & \multicolumn{5}{l}{0.152 / 0.316}                                                                                                                             \\ \hline
\end{tabular}
\end{table*}

\paragraph{Model for subject profiles} The first regression is defined to investigate the behavior of the synopses scores with age, educational level, language and the writer in interaction with the aspects. The resulting model revealed a significant main effect of the writer [$F_{5,7419}=117.23, p < .0001$], and no significant effects of age [$F_{1,10}=1.14, p=.31$], educational background [$F_{2,10}=.767, ns$] and language [$F_{7,10}=.726, ns$]. Therefore, we removed these three slopes from the second model and we focused on the main effect of the experiment: the writer effect and its interaction with the aspects.

\paragraph{Model to verify SLM-human differences} Table \ref{tab:model} shows the estimates and intercept for the second model. We considered fixed effects of writer (SLM or human) and its interactions between the writer and the aspects. We set the creativity aspect as origin for the regression. Taking this reference, the intercept for the model is 1.47 [$95\%\ CI\ 1.23 - 1.70,\ t(40)=,\ p<0.001$]. In general, the answers to the rest of the questions are significant: readability [$95\%\ CI\ 0.94 - 1.1,\ t(7392)=16.70,\ p<0.001$], understandability [$95\%\ CI\ 0.57 - 0.8,\ t(7392)=10.82,\ p<0.001$], relevance [$95\%\ CI\ 0.29 - 0.54,\ t(7392)=6.50,\ p<0.001$] and informativity [$95\%\ CI\ 0.37 - 0.62,\ t(7392)=7.69,\ p<0.001$] questions obtain $1.07$, $0.69$, $0.42$ and $0.49$ over the intercept, respectively. The aspect of attractiveness gets a $-0.25$ under the intercept [$95\%\ CI\ -0.38 - -0.13,\ t(7392)=-3.92,\ p<0.001$]. 

In particular, the model estimated readability, understandability, relevance, informativity and attractiveness of SLM synopses ($SLM \times aspects$ interactions) to be significantly higher than for humans, except in creativity. That is, placing creativity at the origin of the regression, there is no SLM term that differs significantly from the intercept, but the model's interactions with the other aspects are significant.

With respect to random effects, the variability of the intercept across subjects and synopses are $0.21$ and $0.18$, respectively. The interactions $subject \times SLM$ and  $synopsis \times SLM$ in the random effects capture how much subjects’ and synopsis effects deviate from the population of SLM; they are not too large considering that we are on a scale of 0 to 4.

In conclusion, the results of the first linear mixed effect model show that subject attributes (age, educational background, and language) do not have a significant impact in the responses. The second model confirms the significance results obtained with the Wilcoxon signed-rank test in the body of the article.


\subsubsection{Common language effect size}

Table \ref{tab:effect-size} shows the common language effect \citep{vargha2000critique} size for each aspect (the main experiment is reported in the leftmost column). As we can see in the table, the probability that a randomly selected synopsis from the SLM is better than a randomly sampled synopsis from humans is $0.69$.

Examining individual quality aspects, the largest effect size is in grammaticality. The proportion of pairs where the grammar of the SLM is better than the grammar of the human is $0.76$. This confirms the statement of \citet{see_massively_2019}, who claims that current language models (even the small ones) have enough capacity to effectively replicate the distribution of human language. 

\begin{table}
\scriptsize
\centering
\caption{Common Language Effect Size between SLM scores and human scores for each experiment. It is the proportion of pairs (human / SLM-made synopses) where the SLM score is higher than the human score. In other words, the probability that, for a randomly selected title, its SLM synopsis is better than its human synopsis. (* denotes significant difference: **** for $p < .0001$, *** for $p < .001$, ** for $p < .01$, and * for $p < .05$ )}\label{tab:effect-size}

\begin{tabular}{@{}l|ccc@{}}
\toprule
                                      & \multicolumn{1}{l}{\textbf{Main}} & \multicolumn{1}{l}{\textbf{Revealed}} & \multicolumn{1}{l}{\textbf{AI only}} \\ \midrule
\textit{readability}                  & \textbf{0.76****}                               &  \textbf{0.71***}                  &  \textbf{0.77****}                                  \\
\textit{understandability}            & \textbf{0.65**}                               & 0.60                            &  \textbf{0.65**}                               \\
\textit{relevance}                    & \textbf{0.66**}                               & \textbf{0.67***}                   & \textbf{0.70***}                                   \\
\textit{informativity}                & \textbf{0.61**}                               & 0.57                            &  0.57                                  \\
\textit{attractiveness}               & \textbf{0.64**}                               & 0.49                            &  0.59                                  \\
\textit{creativity}                   & 0.47                                        & 0.45                            &  0.47                                  \\ \midrule
\multicolumn{1}{c|}{\textbf{Overall Quality}} & \textbf{0.69}                      & \textbf{0.63}                   & \textbf{0.67}                                    \\ \bottomrule
\end{tabular}
\end{table}

\subsection{Analysis of Outliers}

We have carried out a detailed study of outliers in our dataset, in order to ensure that our results are not biased by the behavior of particular readers or particular synopses that deviate substantially from the others.

\subsubsection{Reader Outliers}
 
 We used four tests to identify outlier readers:

\begin{itemize}
    \item Based on percentiles: we select as outliers those readers whose replies, 50\% of the times or more, are out of the range defined by the percentile 5 and percentile 95.
\item Based on standard deviation: if the difference between the assessor response and the mean of a question in a certain quiz is greater than twice the standard deviation, we consider this reply as outlier. If more than 50\% of the answers of a reader are outliers, the assessor is an outlier.
\item Based on frequency: If half or more of the replies of a reader are the same, it is considered outlier (for example, 50\% of the times, someone replies to all questions with a 2).
\item Based on completion time: the average speed of reading in English is 300 words per minute \citep{brysbaert2019many}. The length of the quizzes is 3400 words, which gives 16.5 minutes. If we consider that replying to a question takes at least three seconds, we get 18 minutes, for a total of 34.5 minutes to complete the quiz. Every reader that completes the quiz in less time is considered an outlier. 
  
\end{itemize}

Using these criteria, seven readers where marked as outliers. Eliminating them does not change the results significantly. In fact, it increases the difference in favor of the SLM as Table \ref{tab:outliers} shows. Therefore, given the difficulty of differentiating between a true outlier and a random outlier, we decided to keep all participants for our analysis.

\begin{table}
\caption{Means for each experiment removing outliers.}
\scriptsize
\label{tab:outliers}
\centering
\setlength{\tabcolsep}{10pt} 
\begin{tabular}{llll}
\hline
                    & \multicolumn{3}{c}{\textbf{Main}}                                                                        \\ \hline
                    & \textit{\shortstack{All \\ participants}} & \textit{\shortstack{Without \\ outliers}} & \textit{\shortstack{Without the \\ fastest readers}} \\ \hline
SLM             & 2.1105                         & 2.0616                    & 2.1174                             \\
human               & 1.8533                         & 1.7577                    & 1.8455                             \\
difference          & \textbf{0.2572}                & 0.3039                    & 0.2719                             \\ \hline
                    & \multicolumn{3}{c}{\textbf{Revealed}}                                                                    \\ \hline
                    & \textit{\shortstack{All \\ participants}} & \textit{\shortstack{Without \\ outliers}} & \textit{\shortstack{Without the \\ fastest readers}} \\  \hline
SLM             & 1.9825                         & 1.9981                    & 1.9981                             \\
human               & 1.8750                         & 1.8796                    & 1.8796                             \\
difference          & \textbf{0.1075}                & 0.1184                    & 0.1184                             \\ \hline
                    & \multicolumn{3}{c}{\textbf{SLM only}}                                                           \\ \hline
                    & \textit{\shortstack{All \\ participants}} & \textit{\shortstack{Without \\ outliers}} & \textit{\shortstack{Without the \\ fastest readers}} \\  \hline
SLM             & 2.1460                         & 2.1326                    & 2.0946                             \\
human               & 1.9301                         & 1.9108                    & 1.8666                             \\
difference          & \textbf{0.2159}                & 0.2218                    & 0.2280                             \\ \hline
\end{tabular}
\end{table}

\subsubsection{Synopses Outliers}

During the qualitative evaluation shown below, we detected two human synopses that seemingly contained subtle human errors. Removing them makes the average quality of human synopsis grow. However, the mean of SLM-generated synopses remains higher even removing these two faulty human synopses.

\section{Linguistic Analysis BART-large vs. GPT 3.5 and GPT-4o}
\label{appx:linguistic}
Table \ref{tab:qualitative-extended} shows more details from the qualitative analysis carried out.

\begin{table*}[t!]
\small

\caption{Frequencies of the qualitative salient linguistic properties of AI-generated synopses. For the formulaic phrases, we report the percentage of synopses in which they appear and the total number of appearances in parentheses. Note that BART is a small size transformer fine-tuned for the synopsis generation task, while GPT-3.5 4o are a large, State-of-the-Art models generating synopses in zero-shot mode. *The average of the 25\% longer synopses is 141,8 words/text for BART, 133,9 words/text for GPT-3.5 and 89,9 words/text for GPT-4o.}
\label{tab:qualitative-extended}
\centering
\begin{tabular}{llrrr}
& & BART (SLM) & GPT-3.5 & GPT-3.5 4.0 \\ 
 & & (fine-tuned) & (zero-shot)  & (zero-shot) \\ 
\hline
{\bf Formulaic Phrases}     & With at least one clich\'e phrase &  83,3\% (50) & {\bf 100\% (60)}  & 26,7\% (16) \\ 
                            & With two or more clich\'e phrases & {\bf 46,7\% (28)} &  25,0\% (15) & 0\% \\ 
        & Collocation ``car accident''  &  {\bf 10,0\% (6)} & 0\%  & 0\% \\
        & Collocation ``young man''  & {\bf 26,7\% (16)} &  8,3\% (5)  & 0\% \\         
        & Collocation ``young woman''  &  {\bf 28,3\% (19)} & 18,3\% (11)  & 1,7\% (1) \\ 
        
        & Collocation ``life is turned upside down''  & 1,7\% (1) & 0\%  & {\bf 5,0\% (3)} \\
        & Collocation ``widowed mother''  &  {\bf 8,4\% (5)} & 0\%  & 0\% \\     
        & Collocation ``south of France''  & {\bf 8,4\% (5)} & 0\%  & 0\% \\    
        & Collocation ``successful businessman''  &  {\bf 8,4\% (5)} & 1,7\% (1)  & 0\% \\
        & Collocation ``time is running out''  & 0\% & 0\%  &  {\bf 5,0\% (3)} \\
        & Collocation ``true meaning''  & 0\%  &  {\bf 16,7\% (10)}  & 0\% \\
        & With ``adventure''& 0\%       &  {\bf 23,3\% (18)}        & 18,3\% (12) \\            
        & With ``ancient''  & 0\%       & 3,3\% (3)                 & {\bf 28,3\% (17)} \\     
        & With ``Clara''    & 0\%       & 0\%                       & {\bf 6,7\% (9)} \\        
        & With ``dark''     & 1,7\% (1) & 11,7\% {\bf (13)}         & {\bf 16,7\%} (11) \\                    
        & With ``Emma''     & 0         & 3,3\% (7)                 & {\bf 10,0\% (12)} \\           
        & With ``family''   & {\bf 26,7\%} (27)     &  {\bf 26,7\% (38)}   &  21,7\% (22) \\ 
        & With ``father''   & {\bf 30,0\% (42)} & 0\%         & 5,0\% (3) \\ 
        & With ``heartwarming''  & 0\% & {\bf 20,0\% (12)}   & 11,7\% (7) \\ 
        & With ``hidden''   & 0\% & 3,3\% (2)                & {\bf 36,7\% (23)} \\ 
        & With ``Jack''     & 1,7\% (1) & {\bf 10,0\% (15)}         & 3,3\% (3) \\ 
        & With ``journey''  & 0\% & {\bf 30,0\% (22)}        & 28,3\% (17) \\ 
        & With ``life''     & 13,3\% (8) & {\bf 40,0\% (37)}        & 28,3\% (24) \\ 
        & With ``Lily''  & 0\% & 0\%                    & {\bf 8,3\% (10)} \\ 
        & With ``love''  & 35,0\% (23) & {\bf 40,0\% (39)}        & 26,7\% (25) \\                     
        & With ``marr[ies/y/age]''  & {\bf 21,7\% (20)} & 3,3\% (5)  & 3,3\% (4) \\ 
        & With ``mother''  &  {\bf 23,3\% (14)} & 5\% (5)      & 0\% \\ 
        & With ``myster[y/ies/ious]''  & 0\% & 13,3\% (8)   & {\bf 33,3\% (22)} \\       
        & With ``secret[s]''  & 0\% & 20,0\% (12)            & {\bf 55,0\% (35)} \\ 
        & With ``true''  & 0\% &  {\bf 41,7\% (29)}          & 21,7\% (13) \\                     
        & With ``truth''  & 1,7\% (1) &  13,3\% {\bf(13)}    & {\bf 18,3\%} (12) \\ 
        & With ``ultimately''  & 0\% & {\bf 33,3\% (21)}     & 6,7\% (4) \\ 
        & With ``uncover''  & 0\% & 15,0\% (12)             & {\bf 36,7\% (22)} \\ 
        & With ``unexpect[ed/edly]''  & 0\% & 8,3\% (5)     & {\bf 25,0\% (16)} \\     
        & With ``war''  &  {\bf 16,7\%} (12) &  3,3\% {\bf (18)}        & 5,0\% (15) \\                     
                    \hline
{\bf External Coherence}  & Overall coherence with external facts & 86,7\% & 91,7\%  & {\bf 100\%} \\ 
                    & When referring to a place and a date & 62,5\% & 60,0\% & {\bf 100\%} \\ 
                    & When referring to a specific historical event & 50,0\%  & 30,0\% & {\bf 100\%} \\ \hline
{\bf Internal Coherence}  & Overall & 68,3\% & 95,0\% & {\bf 100\%} \\ 
                    & Of the top 25\% longer synopses* & 33,3\% &  100\% & {\bf 100\%} \\ \hline
{\bf Surprising Associations} &  & {\bf 15,0\%} & 3,3\%  & 1,6\% \\  \hline
{\bf Recurrent Themes}   & Adventure         & 3,3\%       & 20,0\%   & {\bf 25,0\%} \\
                         & Family            & {\bf 13,3\%}      & {\bf 13,3\%}   & 8,3\% \\
                         & Friendship        & {\bf 11,7\%}      & 5,0\%    & 3,3\% \\                    
                         & Love              & {\bf 36,7\%}      & 18,3\%   & 11,7\% \\     
                         & Mystery \& Crime          & 18,3\%      & 16,7\%   & {\bf 28,3\%} \\  
                         & Personal Growth   & 3,3\%       & {\bf 23,3\%}   & 20,0\% \\   
                         & War               & {\bf 13,3\%}      & 3,3\%    & 3,3\%  \\                           
                         
\hline
\end{tabular}
\end{table*}

\section{Examples of Qualitative Analysis of Synopses}
\label{appx:examples}

This appendix includes several sample texts that illustrate the textual characteristics taken into account in the qualitative analysis of the synopses.

\subsection{Examples of Coherence with External Facts}

Example \ref{ex2} sets the English Civil War in the late 1920s, but it actually occurred long before; and Example \ref{ex3} sets the action in the aftermath of $11^{th}$ September “on the other side of the Iron Curtain”, but the Iron Curtain had already fallen long before.

\begin{subexamples}[preamble={ \textit{\added{English: An Autumn In London}.}}]\label{ex2}
		\item English: An Autumn in London tells the story of a group of English students who are invited to spend a weekend at a country house in the English countryside. The film is set in the late 1920s and early 1930s, during the English Civil War. The main character is a young Englishman who has been sent to London to complete his education. He meets a young woman who is also studying in London and falls in love with her. However, she is engaged to be married to the son of a local aristocrat. {\it(SLM)}\\
		\item\label{ex2:human} 
		\added{The lives of five characters from diverse backgrounds whose passage to the U.K. and its aftermath are dealt with in the film. Jayasurya plays Shankaran, a Kathakali artiste-turned-waiter who is an illegal immigrant. Nivin Pauly dons the role of Sibin, an IT executive with a roving eye. Mukesh plays Joy, a middle-class corner store owner with an extended family in London and all its concomitant problems and advantages. Nadia Moidu plays Saraswathy, a Tamil Brahmin. Married to a doctor, she has been in the U.K, for more than 20 years. Remya Nambeesan plays Gauri, a young married woman from a rustic background who arrives in London.} \added{{\it(Human)}}\\
	\end{subexamples}
    
     \begin{subexamples}[preamble={\textit{\added{The Other Side}}}]\label{ex3}
		\item \added{The Other Side tells the story of a group of friends on the other side of the Iron Curtain who are forced to choose between their loyalty to one side or the other. The film is set in the aftermath of the September 11th attacks.} \added{{\it(SLM)}}
		\\
		\item \added{Max Mackenzie, an over-achieving science student, receives a scholarship by Marston I and Marston II to study fish on the remote Destiny Island. On the island she meets an array of odd citizens leading her to discover a tragedy that happened on the island years prior. She also falls in love with Sean Splinter , the mayor's son.} \added{{\it(Human)}} \\
	\end{subexamples}

GPT-4o is especially good with internal consistency. For example, if the film takes place in Mexico the main family is called Sanchez, or as seen in Example \ref{ExConventChatGPT4} with various Italian and clerical references.

\begin{subexamples}[preamble={\textit{\added{The Convent}}}]\label{ExConventChatGPT4}
		\item\label{Call:model} In the remote hills of northern Italy lies a forgotten convent, shrouded in mystery and silence. 'The Convent' follows Sister Isabella as she uncovers ancient secrets sealed within its stone walls. When strange occurrences plague the nuns, she must confront the sinister force lurking in the shadows. Blending Gothic horror with a tale of redemption, this thriller explores the delicate balance between faith and the supernatural. Florence 1912 Italy. (GPT 4o).
	\end{subexamples}

\subsection{Examples of Internal Coherence} Example \ref{ex:Miniver} stated that Miniver's father died in a car accident but, after that, he gets married.

 \begin{subexamples}[preamble={\textit{\added{The Miniver Story}}}]\label{ex:Miniver}
		\item \added{The Miniver Story tells the story of a young boy named Miniver, who was abandoned by his parents when he was a baby. He grows up to become a successful businessman, but his father dies in a car accident, leaving Miniver to fend for himself on his own. Miniver's mother dies, leaving him with his father's will, which states that he will inherit his mother's fortune if he marries a woman of his choice. He marries the daughter of a wealthy family, but after the marriage, he finds out that she is in love with another man. He leaves her and moves to the United States, where he becomes a successful business man. Years later, Miniver returns to his home town and finds that his mother has died, and his father has remarried. He decides to return to his old life, but is unable to forget his mother. He meets a young woman named Mimi, and they fall in love, but Mimi's father disapproves of their relationship, and forbids them to see each other again. He sends Miniver away to live with his aunt and uncle, and Mimi becomes pregnant, but dies shortly after giving birth to a baby boy.} \added{{\it(SLM)}}
		\\
		\item \added{The story, told partly in flashback and narrated by Clem Miniver, commences on VE Day as Clem and Judy return home from war service and Toby returns from a foster family in America. Judy, a corporal driver, is loved by Tom Foley, a captain in the Royal Engineers, but she is besotted with a general by the time The Miniver Story was produced in 1950.} \added{{\it(Human)}}
		\\
	\end{subexamples}

\subsection{Examples of Surprising Associations} Sometimes, AI generated synopsis make surprising or even humorous associations, which could be seen as creative, as in Examples \ref{convent} and \ref{pickpocket}.

 \begin{subexamples}[preamble={\textit{\added{The Convent}}}]\label{convent}
		\item\label{convent:model} \added{The Convent of Notre Dame de Namur tells the story of a young girl who is sent to a convent after the death of her parents. She is placed under the care of Sister Catherine, a strict disciplinarian, and is forced to live in the convent for the rest of her life. She soon finds out that the convent is not a convent at all, but rather an institution for the criminally insane, and she soon finds herself falling in love with Sister Catherine.} \added{{\it(SLM)}} \\
		\item\label{convent:human} \added{The journey of Michael Padovic, a professor from the U.S.A, who arrives with his wife, Helene, at a Portuguese convent where he expects to find the documents needed to prove his theory: Shakespeare was born in Spain, not in England, and was Jewish. The main characters arrive at the convent where they are greeted by an enigmatic stranger who refers to himself as Baltar; he is the keeper of the convent. Other characters include a fisherman, the beautiful bookkeeper who is attracted to Dr. Padovic's work, and an elderly man who goes by the name of Balthazar.} \added{{\it(Human)}} \\
	\end{subexamples}
    
    \begin{subexamples}[preamble={\textit{\added{Beware Of Pickpockets}}}]\label{pickpocket}
		\item \added{The film tells the story of a group of pickpockets who are on the run from the police. The film is set in a small town in the south of France during the early 1970s. The main character is a pickpocket named Jean-Claude, who lives with his widowed mother, who works as a waitress in a restaurant, and his younger brother, Jean-Michel, who is a student at the local high school. The two brothers are in love with the same girl, Marie, but she is engaged to the son of a wealthy family, and the father does not approve of their relationship because he fears that he will lose his job if he marries her. Jean and Michel's relationship is further complicated by the fact that Marie's father is a member of the pickpocket gang, and he is also the leader of the gang.} \added{{\it(SLM)}}
		\\
		\item \added{Righteous officer Big Nose Pau is ordered to arrest pickpocket Extra Hand. However, each time he was caught, he was released due to a lack of evidence. Although Extra Hand is a tricky man, he is actually a Robin Hood like pickpocket who steals dirty money from the rich to raise seven orphaned children that he adopted. Extra Hand plans to save money to build an orphanage for the children to live in. Later, Extra Hand learns that his rival, Dog Lice, has robbed many jewels and therefore, Extra Hand planned to steal them from him. After a major battle with Dog Lice and his gang, Extra Hand was caught by Pau. Although Pau was hesitant to arrest him, Extra Hand decides not to give a tough job for Big Nose and surrenders to him. However, seeing how Extra Hands is doing all this for the orphans, Pau decides to let him go. {\it(Human)}}
		\\
	\end{subexamples}

\bibliography{references}

\end{document}